\documentclass[11pt]{article}
\usepackage{eacl2017}
\usepackage{times}
\usepackage{url}
\usepackage{latexsym}
\usepackage[english]{babel}
\usepackage{amsmath}
\usepackage{amssymb}
\usepackage[usenames,dvipsnames]{xcolor}
\usepackage{tabularx, multirow}
\usepackage{graphicx}
\usepackage{comment}

\newcommand{\yg}{{\bf y}_{\textrm{g}}}
\newcommand{\pc}{{\bf f}}
\newcommand{\fcorr}{{f}_{\textrm{cor}}}
\newcommand{\fwrong}{{f}_{\textrm{wrong}}}
\newcommand{\fstat}{{f}_{\textrm{stat}}}
\newcommand{\yref}{{\bf y}_{\textrm{ref}}}
\newcommand{\ypred}{{\bf y}_{\textrm{pred}}}
\newcommand{\X}{{\mathcal{X}}}
\newcommand{\Y}{{\mathcal{Y}}}
\newcommand{\x}{{\bf x}}
\newcommand{\y}{{\bf y}}
\newcommand{\s}{{\bf s}}
\newcommand{\att}{{\bf a}}
\newcommand{\sconf}{{\bf s}_{\textrm{conf}}}
\newcommand{\sproduct}{{\bf s}_{\textrm{pr}}}
\newcommand{\sclassifier}{{\bf s}_{\textrm{cl}}}
\newcommand{\F}{{\bf F}}
\newcommand{\Fdual}{{\bf F}_{\textrm{dual}}}
\newcommand{\Fxyg}{\F\left(\x, \yg \right)}
\newcommand{\embS}{{\bf S}}
\newcommand{\embT}{{\bf T}}

\newcommand{\nn}{{\textrm{nn}}}
\newcommand{\Prob}{\textrm{P}}

\newcommand{\err}[1]{{\bf \color{BrickRed} #1}}
\newcommand{\corr}[1]{{\bf \color{OliveGreen} #1}}
\newcommand{\mix}[1]{{\bf \color{BurntOrange} #1}}

\def\textit#1{{\it #1}}
\def\textbf#1{{\bf #1}}
\def\textsl#1{{\sl #1}}
\def\texttt#1{{\tt #1}}

\newcommand{\citet}{\newcite}
\newcommand{\citep}{\cite}

\definecolor{dual}{HTML}{D95F02}
\definecolor{single}{HTML}{66A61E}
\definecolor{fulloracle}{HTML}{E7298A}
\definecolor{partialoracle}{HTML}{A6761D}

\eaclfinalcopy 

\title{Iterative Refinement for Machine Translation}

\author{Roman Novak \\
  Ecole Polytechnique\thanks{~~Roman was interning at Facebook for this work.} \\
  \\\And
  Michael Auli \\
  Facebook AI Research \\
  \\\And
  David Grangier\\
  Facebook AI Research \\
}

\date{}

\begin{document}
\maketitle

\begin{abstract}
Existing machine translation decoding algorithms generate 
translations in a strictly monotonic fashion and never 
revisit previous decisions. 
As a result, earlier mistakes cannot be corrected at a later 
stage.
In this paper, we present a translation scheme that starts 
from an initial guess and then makes iterative improvements 
that may revisit previous decisions.
We parameterize our model as a convolutional neural network 
that predicts discrete substitutions to an existing translation
based on an attention mechanism over both the source sentence 
as well as the current translation output.
By making less than one modification per sentence, we improve 
the output of a phrase-based translation system by up to
0.4 BLEU on WMT15 German-English translation.
\end{abstract}

\section{Introduction}\label{sec:Introduction}

Existing decoding schemes for translation generate outputs
either left-to-right, such as for phrase-based or neural
translation models, or bottom-up as in syntactic models 
\cite{koehn:2003:naacl,galley:2004:naacl,bahdanau:2015:iclr}.
All decoding algorithms for those models make decisions
which cannot be revisited at a later stage, such as when the
model discovers that it made an error earlier on.

On the other hand, humans generate all but the simplest 
translations by conceiving a rough draft of the solution 
and then iteratively improving it until it is deemed complete.
The translator may modify a clause she tackled earlier 
at any point and make arbitrary modifications to improve
the translation.

It can be argued that beam search allows to recover from mistakes,
simply by providing alternative translations. 
However, reasonable beam sizes encode only a small number of
binary decisions. 
A beam of size $50$ contains fewer than six binary decisions,
all of which frequently share the same prefix
\cite{huang:2008:thesis}.\footnote{$2^5 = 32 < 50 < 2^6 = 64$}

In this paper, we present models that tackle translation
similar to humans. The model iteratively edits
the target sentence until it cannot improve it further. 
As a preliminary study, we address the problem of finding mistakes 
in an existing translation via a simple classifier that predicts 
if a word in a translation is correct (\textsection\ref{sec:Error Detection}).
Next, we model word substitutions for an existing translation
via a convolutional neural network that attends to the source
when suggesting substitutions 
(\textsection\ref{ssec:Attention-based Model}).
Finally, we devise a model that attends both to the source 
as well as to the existing translation
(\textsection\ref{ssec:Dual Attention Model}).
We repeatedly apply the models to their own output by determining 
the best substitution for each word in the previous translation and 
then choosing either one or zero substitutions for each sentence. 
For the latter we consider various heuristics as well as a
classifier-based selection method 
(\textsection\ref{ssec:Refinement Strategy}).

Our results demonstrate that we can improve the output of a 
phrase-based translation system on WMT15 German-English data
by up to 0.4 BLEU \cite{Papineni:2002:BMA:1073083.1073135} by 
making on average only $0.6$ substitutions 
per sentence (\textsection\ref{sec:Experiments}).

Our approach differs from automatic post-editing
since it does not require post-edited text which is a scarce
resource \cite{simard:2007:naacl,bojar:2016:wmt}. 
For our first model (\textsection\ref{ssec:Attention-based Model}) 
we merely require parallel text and for our second model 
(\textsection\ref{ssec:Dual Attention Model}) the output 
of a baseline translation system.

\section{Detecting Errors}\label{sec:Error Detection}

Before correcting errors we consider the task of detecting mistakes in the output of an existing translation system. 

	In the following, we use lowercase boldface for vectors (e.g. $\x$), uppercase boldface for matrices (e.g. ${\bf F}$) and calligraphy for sets (e.g. $\X$). 
    We use superscripts for indexing or slicing, e.g., $\x^i$, $\F^{i,j}$, $\F^i=(\F^{i,1},\dots,\F^{i, |\F^i|})$.
	We further denote $\x$ as the source sentence, $\yg$ as the guess translation from which we start and which was produced by a phrase-based translation system (\textsection\ref{sec:Experimental Setup}), and $\yref$ as the reference translation. Sentences are vectors of indices indicating entries in a source vocabulary $\X$ or a target vocabulary $\Y$. For example,
	\mbox{$\x =(\x^1,\dots,\x^{|\x|}) \in \mathcal{X}^{|\x|}$} with $\X = \left\{1, \dots, |\X|\right\}$. We omit biases of linear layers to simplify the notation.

Error detection focuses on word-level accuracy, i.e., we predict for each token in a given translation if it is present in the reference or not. This metric ignores word order, however, we hope that performance on this simple task provides us with a sense of how difficult it will be to modify translations to a positive effect.
    A token $\yg^i$ in the candidate translation $\yg$ is deemed correct iff it is present in the reference translation: $\yg^i \in \yref$. We build a neural network $\pc$ to predict correctness of each token in $\yg$ given the source sentence $\x$:
$$\pc(\x, \yg) \in \left[0;1\right]^{|\yg |},$$
where $\pc(\x, \yg)^i$ estimates $\Prob\left(\yg^i \in \yref\right).$ 
		
\textbf{Architecture.} We use an architecture similar to the word alignment model of \citet{Align}. The source and the target sequences are embedded via a 
lookup table that replace each word type with a learned vector. The resulting vector sequences are then processed by
alternating convolutions and non-linearities. This results in a vector $\embS\left(\x\right)^i$ representing each position $i$ in the source $\x$ and a vector $\embT\left(\yg\right)^j$ representing each position $j$ in the target $\yg$. These vectors are then compared via a dot product. Our prediction estimates the probability of a target word being correct as the largest dot product between any source word and the guess word.
We apply the logistic function $\sigma$ to this score,
$$\pc(\x,\yg)^i = \sigma\left(\max_{1\leqslant j \leqslant |\x|}\left[\embS(\x)\embT(\yg)^T\right]^{j, i}\right).$$
		
\textbf{Training.} At training time we minimize the cross-entropy loss, with the binary supervision $1$ for $\yg^i \in \yref$, 
$0$ otherwise.

\textbf{Testing.} At test time we threshold the model prediction $\pc(\x,\yg)^i$ to detect mistakes. We compare the performance of our network to the following baselines:
		\begin{enumerate}
			\item Predicting that all candidate words are always correct $\fcorr \equiv {\bf 1}$, or always incorrect $\fwrong \equiv {\bf 0}$.
        \item The prior probability of a word being correct based on the training data $\fstat(y) = \left(\Prob\left[y \in \yref \,|\, y \in \yg\right] > 0.5\right)$.
		\end{enumerate}
		
We report word-level accuracy metrics in Table~\ref{tab:err}. While the model significantly improves over the baselines, the probability of correctly labeling a word as a mistake remains low ($62.71\%$). The task
of predicting mistakes is not easy as previously shown in confidence estimation \cite{blatz:2004:coling,ueffing:2007:cl}. Also, one should bear in mind that this task cannot be solved with $100\%$ accuracy since a sentence can be correctly in multiple different ways and we only have a single reference translation. In our case, our final refinement objective might be easier than error detection as we do not need to detect all errors. We need to identify some of the locations where a substitution could improve BLEU. At the same time, our strategy should also suggest these substitutions. This is the objective of the model introduced in the next section.

		\begin{table}[t]
			\centering
			\begin{tabular}{|l|r|r|r|r|}
				\hline
				\multicolumn{1}{|c|}{Metric (\%)} & $\fcorr$ & $\fwrong$ & $\fstat$ & $\pc$ \\ \hline
				Accuracy          & 68.0           & 32.0			     	& 71.3               & \textbf{76.0}      \\ \hline
				Recall            & 0.00               & \textbf{100.00}		    & 36.0               & 61.3      \\ \hline
				Precision         & \textbf{100.0}    & 32.0					& 58.4               & 62.7      			\\ \hline
				F1                & 0.00 				& 48.4                	& 44.5               & \textbf{62.0}      \\ \hline
		\end{tabular}%
			\caption{Accuracy of the error detection model $\pc$ compared to baselines on the concatenation of the WMT test sets from $2008$ to $2015$. For precision, recall and F1 we consider a positive prediction as labeling a word as a mistake. Baseline $\fcorr$ labels all words as correct, $\fwrong$ labels all words as incorrect, $\fstat$ labels a word from $\yg$ based on the prior probability estimated on the training data.}
			\label{tab:err}
		\end{table}

\section{Attention-based Model}\label{ssec:Attention-based Model}

We introduce a model to predict modifications to a translation which can be trained on bilingual text. In \textsection\ref{ssec:Refinement Strategy} we discuss strategies to iteratively apply this model to its own output in order to improve a translation.

Our model $\F$ takes as input a source sentence $\x$ and a target sentence
$\y$, and outputs a distribution over the vocabulary for each target position,
$$ 
\F(\x, \y) \in \left[0, 1\right]^{|\y| \times |\Y| }.
$$
For each position $i$ and any word $j \in \Y$, $\F(\x, \y)^{i, j}$ estimates $\Prob(\y^i = j\,|\, \x, \y^{-i})$, the probability of word $j$ being at position $i$ given the
source and the target context $\y^{-i} = \left(\y^1,\dots,\y^{i-1},\y^{i+1},\dots,\y^{|\y|}\right)$ surrounding $i$. In other words, we 
learn a non-causal language model \cite{Bengio:2003:NPL:944919.944966} which is also conditioned on the source $\x$.
    
\textbf{Architecture.} We rely on a convolutional model with attention.
The source sentence is embedded into distributional space via a lookup table,
followed by convolutions and non-linearities. The target sentence is also embedded
in distributional space via a lookup table, followed by a single convolution and 
a succession of linear layers and non-linearities. The target convolution weights are zeroed at the center so that the model does not have access to the center word. This means that the model observes a fixed size context of length $2k$ for any target position $i$,
$
\y^{-i | k} = \left(\y^{i - k}, \dots,\y^{i - 1},\y^{i + 1},\dots,\y^{i+k}\right)
$
where $2k + 1$ refers to the convolution kernel width. These operations
result in a vector $\embS^{j}$ representing each position $j$ in the source sentence $\x$ and a vector $\embT^{i}$ representing each target context $\y^{-i | k}$.

Given a target position $i$, an attention module then takes as input these representation and outputs a weight for each target position
$$
\alpha(i, j) = 
\frac{\exp\left(\embS^{j} \cdot \embT^{i}\right)}
{\sum_{j' = 1}^{|\x|}\exp\left(\embS^{j'} \cdot \embT^{i}\right)}.
$$
These weights correspond to dot-product attention scores~\cite{conf/emnlp/LuongPM15,rush:2015:emnlp}. 
The attention weights allow to compute a source summary specific to each target context through a weighted sum,
$$
\att\left(\y^{-i|k}, \x\right) = \sum_{j=1}^{|\x|} \alpha(i, j) ~ \embS^{j}
$$
Finally, this summary $\att(\y^{-i|k}, \x)$ is concatenated 
with the embedding of the target context ${\y}^{-i|k}$
obtained from the target lookup table,
$$
{\bf L}\left({\y}^{-i|k}\right) = 
\left\{ 
{\bf L}^j, j \in  \y^{-i | k} 
\right\} 
$$
and a multilayer perceptron followed by a softmax computes
$\F(\x,\y)^i$ from $\att(\y^{-i|k}, \x),~{\bf L}({\y}^{-i|k})$.
Note that we could alternatively use $\embT^{i}$ instead of
${\bf L}({\y}^{-i|k})$ but our preliminary validation
experiments showed better result with the lookup table output.

\textbf{Training.} The model is trained to maximize the 
(log) likelihood of the pairs $(\x, \yref)$ from the training set.
        
\textbf{Testing.} At test time the model is given $(\x, \yg)$,
i.e., the source and the guess sentence. Similar to maximum
likelihood training for left-to-right translation systems~\cite{bahdanau:2015:iclr}, the model is therefore 
not exposed to the same type of context in training (reference contexts from $\yref$) and testing (guess contexts from $\yg$).

\textbf{Discussion.} Our model is similar to the attention-based translation approach of \citet{bahdanau:2015:iclr}. In addition
to using convolutions, the main
difference is that we have access to both left and right target 
context $\y^{-i|k}$ since we start from an initial guess translation. 
Right target words are of course good predictors of the previous
word. For instance, an early validation experiment with the setup from \textsection\ref{sec:Experimental Setup} showed a perplexity of
$5.4$ for this model which compares to $13.9$ with the same model trained with the left context only.
        
\section{Dual Attention Model}\label{ssec:Dual Attention Model}

We introduce a dual attention architecture to also make use of the
guess at training time. This contrasts with the model introduced
in the previous section where the guess is not used during training.
Also, we are free to use the entire guess, including the center word,
compared to the reference where we have to remove the center word.

At training time, the dual attention model takes $3$ inputs, that is, the source, the guess and the reference. At test time, the reference
input is replaced by the guess. Specifically, the model
$$
\Fdual(\x, \yg, \yref) \in \left[0; 1\right]^{|\yref|\times |\Y|}
$$
estimates $\Prob\left(\yref^i \,|\, \x, \yg, \yref^{-i}\right)$
for each position $i$ in the reference sentence.

\textbf{Architecture.} The model builds upon the single attention
model from the previous section by having two attention functions $\att$ 
with distinct parameters. 
The first function $\att_{\rm source}$
takes the source sentence $\x$ and the reference context 
$\yref^{-i}$ to produce the source summary for this context
$
\att_{\rm source}\left(\y^{-i|k}, \x\right)
$
as in the single attention model. The second function $\att_{\rm guess}$
takes the guess sentence $\yg$ and the reference context 
$\yref^{-i}$ and produces a guess summary for this context
$
\att_{\rm guess}\left(\y^{-i|k}, \yg\right).
$
These two summaries are then concatenated with the
lookup representation of the reference context
$
{\bf L}\left({\yref}^{-i|k}\right)
$
and input to a final multilayer perceptron followed by
a softmax. The reference lookup table contains the only
parameters shared by the two attention functions.
		
\textbf{Training.} This model is trained similarly to the single attention model, the only difference being the conditioning on the guess $\yg$. 

\textbf{Testing.} At test time, the reference is unavailable and
we replace $\yref$ with $\yg$, i.e., the model is given $(\x, \yg, \yg^{-i|k})$ to make a prediction at position $i$.
In this case, the distribution shift when going from training to testing 
is less drastic than in \textsection\ref{ssec:Attention-based Model} and 
the model retains access to the whole $\yg$ via attention. 

\textbf{Discussion.} Compared to the single attention model (\textsection\ref{ssec:Attention-based Model}), this model  reduces perplexity from $5.4$ to $4.1$ on our validation set. 
Since the dual attention model can attend to all 
guess words, it can copy any guess word if necessary.  
In our dataset, 68\% of guess words are in the reference and can therefore be copied.
This also means that for the remaining 32\% of reference tokens the
model should not copy.
Instead, the model should propose a substitution by itself (\textsection\ref{sec:Experimental Setup}). During testing, the fact that the guess is input twice $(\x, \yg, \yg^{-i|k})$ means that the guess and the prediction context always match. This makes the model more conservative in its predictions, suggesting tokens from $\yg$ more often than the single attention model. However, as we show in \textsection\ref{sec:Experiments}, this turns out beneficial in our setting.

\section{Iterative Refinement}\label{ssec:Refinement Strategy}

The models in \textsection\ref{ssec:Attention-based Model} and \textsection\ref{ssec:Dual Attention Model} suggest word substitutions for each position in the candidate translation $\yg$ given the current surrounding context. 

Applying a single substitution changes the context of the surrounding words and requires updating the model predictions.
We therefore perform multiple rounds of substitution. At each round, the model computes its predictions, then our refinement
strategy selects a substitution and performs it unless the strategy decides that it can no longer improve the target sentence. This means that the
refinement procedure should be able to (i) prioritize the suggested substitutions, and (ii) decide to stop the iterative process.

We determine the best edit for each position $i$ in $\yg$ by selecting the word with the highest 
probability estimate: 
$$
\ypred^i = \arg\max_{j\in \Y}~\Fxyg^{i, j}.
$$
Then we compute a confidence score in this prediction
$
\s(\yg, \ypred)^i
$
, possibly considering the prediction for the current 
guess word at the same position.

These scores are used to select the next position to edit,
$$
i^\star = \arg\max_i~\s(\yg, \ypred)^i
$$
and to stop the iterative process, i.e.,
when the confidence falls below a validated threshold $t$.
We also limit the number of substitutions to a maximum of $N$.
We consider different heuristics for $\s$,
\begin{itemize}

\item Score positions based on the model confidence in $\ypred^i$,
i.e.,
$$
\sconf(\yg, \ypred)^i = \F(\x, \yg)^{i, \ypred^i}.
$$

\item Look for high confidence in the suggested substitution $\ypred^i$ and low confidence in the current word $\yg^i$:
\begin{multline*}
\sproduct(\yg, \ypred)^i \\= 
\F(\x, \yg)^{i, \ypred^i} 
\times
\left(1 - \F(\x, \yg)^{i, \yg^i}\right).
\end{multline*}
\item Train a simple binary classifier taking as input the score of the best predicted word and the current guess word:
\begin{multline*}
\sclassifier(\yg, \ypred)^i\\=
\nn\left(\log\F(\x, \yg)^{i, \ypred^i}, \log\F(\x, \yg)^{i, \yg^i}\right),
\end{multline*}
where $\nn$ is a 2-layer neural network trained to predict
whether a substitution leads to an increase in BLEU or not.
\end{itemize}
		We compare the above strategies, different score thresholds $t$, and the maximum number of modifications per sentence allowed $N$ in \textsection\ref{sec:Results}.

\section{Experiments \& Results}\label{sec:Experiments}

We first describe our experimental setup and then discuss our results.

\subsection{Experimental Setup}\label{sec:Experimental Setup}

\textbf{Data. }We perform our experiments on the German-to-English WMT15 task \cite{WMT:2015} and benchmark our improvements against the output of a phrase-based translation system (PBMT; Koehn et al. 2007)\nocite{koehn:2007:acl} on this language pair. In principle, our approach may start from any initial guess translation. We chose the output of a phrase-based system because it provides a good starting point that can be computed at high speed. This allows us to quickly generate guess translations for the millions of sentences in our training set.
        
All data was lowercased and numbers were mapped to a single special ``number'' token. Infrequent tokens were mapped to an ``unknown'' token which resulted in dictionaries of $120$K and $170$K words for English and German respectively.
        
For training we used $3.5$M sentence triples (source, reference, and the guess translation output by the PBMT system). A validation set of 180K triples was used for neural network hyper-parameter selection and learning rate scheduling. Finally, two $3$K subsets of the validation set were used to train the classifier discussed in \textsection\ref{ssec:Refinement Strategy} and to select the best model architecture (single vs dual attention) and refinement heuristic.
        
The initial guess translations were generated with phrase-based systems trained on the same training data as our refinement models. We decoded the training data with ten systems, each trained on 90\% of the training data in order to decode the remaining 10\%. This procedure avoids the bias of
generating guess translation with a system that was trained on the same data.
        
\textbf{Implementation.} All models were implemented in Torch \cite{Torch} and trained with stochastic gradient descent to minimize the cross-entropy loss. 
        
For the error detection model in \textsection\ref{sec:Error Detection} we used two temporal convolutions on top of the lookup table, each followed by a $\tanh$ non-linearity to compute ${\embS}(\x)$ and ${\embT}(\yg)$. The output dimensions of each convolution was set to $256$ and the receptive fields spanned 5 words, resulting in final outputs summarizing a context of 9 words.
        
For the single attention model we set the shared context embedding dimension $\dim \embS^j = \dim \embT^i = 512$ and use a context of size $k = 4$ words to the left and to the right, resulting in a window of size 9 for the source and 8 for the target. The final multilayer perceptron has 2 layers with a hidden dimension of 
$512$, see \textsection\ref{ssec:Attention-based Model}).
        
For the dual attention model we used 2-layer context embeddings (a convolution followed by a linear with a tanh in between), each having output dimension $512$, context of size $k = 4$. The final multilayer perceptron has 2 layers with a hidden dimension of 
$1024$, see \textsection\ref{ssec:Dual Attention Model}).
In this setup, we replaced dot-product attention with MLP attention \cite{bahdanau:2015:iclr} as it performed better on the validation set.

All weights were initialized randomly apart from the word embedding layers, which were pre-computed with Hellinger Principal Component Analysis \cite{Lebret_EACL_2014} applied to the bilingual co-occurrence matrix constructed on the training set. The word embedding dimension was set to 256 for both languages and all models.

\subsection{Results}\label{sec:Results}

Table~\ref{tab:val_res} compares BLEU of the single and dual attention models ($\F$ vs $\Fdual$) over the validation set. It reports
the performance for the best threshold  $t \in \{0, 0.1,\dots, 1\}$ and 
the best maximum number of modifications per sentence $N \in \{ 0, 1,\dots,10\}$ for the different refinement heuristics.
	
The best performing configuration is $\Fdual$ with the product-based heuristic $\sproduct$ thresholded at $t = 0.5$ for up to $N = 5$ substitutions. We report test performance of this configuration in table \ref{tab:test_res}. 
Tables \ref{tab:good_examples}, \ref{tab:mixed_examples} and \ref{tab:bad_examples} show examples of system outputs. Overall the system obtains a small but consistent improvement over all the test sets.

	\begin{table}[t]
		\centering
			\begin{tabular}{|c|l|l|l|l|}
			\hline
			\multicolumn{1}{|l|}{Model} & \multicolumn{1}{l|}{Heuristic} & Best $t$     & Best $N$   & BLEU  \\ \hline
            \multicolumn{4}{|l|}{PBMT Baseline} & 30.02 \\ \hline
			\multirow{3}{*}{$\F$}      & $\sconf$                          & 0.8          & 3          & 30.21          \\ \cline{2-5} 
			                            & $\sproduct$                         & 0.7          & 3          & 30.20          \\ \cline{2-5} 
			                            & $\sclassifier$                         & 0.5          & 1          & 30.19          \\ \hline
			\multirow{3}{*}{$\Fdual$}      & $\sconf$                          & 0.6          & 7          & 30.32           \\ \cline{2-5} 
			                            & $\sproduct$                         & \textbf{0.5} & \textbf{5} & \textbf{30.35} \\ \cline{2-5} 
			                            & $\sclassifier$                         & 0.4          & 2          & 30.33          \\ \hline
			\end{tabular}
		\caption{Validation results of different model architectures, substitution heuristics, decision thresholds $t$, and number of maximum allowed modifications $N$. Accuracy is reported on a 3041 sentence subset of the validation set.}
		\label{tab:val_res}
	\end{table}
	
	\begin{table}[t]
		\centering
		\begin{tabular}{|l|l|l|l|}
			\hline
			newstest & PBMT BLEU & Our BLEU       & $\Delta$ \\ \hline
			2008             & 21.29        & \textbf{21.60}  & 0.31         \\ \hline
			2009             & 20.42        & \textbf{20.74} & 0.32         \\ \hline
			2010             & 22.82        & \textbf{23.13} & 0.31         \\ \hline
			2011             & 21.43        & \textbf{21.65} & 0.22         \\ \hline
			2012             & 21.78        & \textbf{22.10}  & 0.32         \\ \hline
			2013             & 24.99        & \textbf{25.37} & 0.38         \\ \hline
			2014             & 22.76        & \textbf{23.07} & 0.31         \\ \hline
			2015             & 24.40         & \textbf{24.80}  & 0.40          \\ \hline
			Mean             & 22.49        & \textbf{22.81} & 0.32         \\ \hline
		\end{tabular}
		\caption{Test accuracy on WMT test sets after applying our refinement procedure.}
		\label{tab:test_res}
	\end{table}

\begin{figure}
		\centering
        \includegraphics[width=0.5\textwidth]{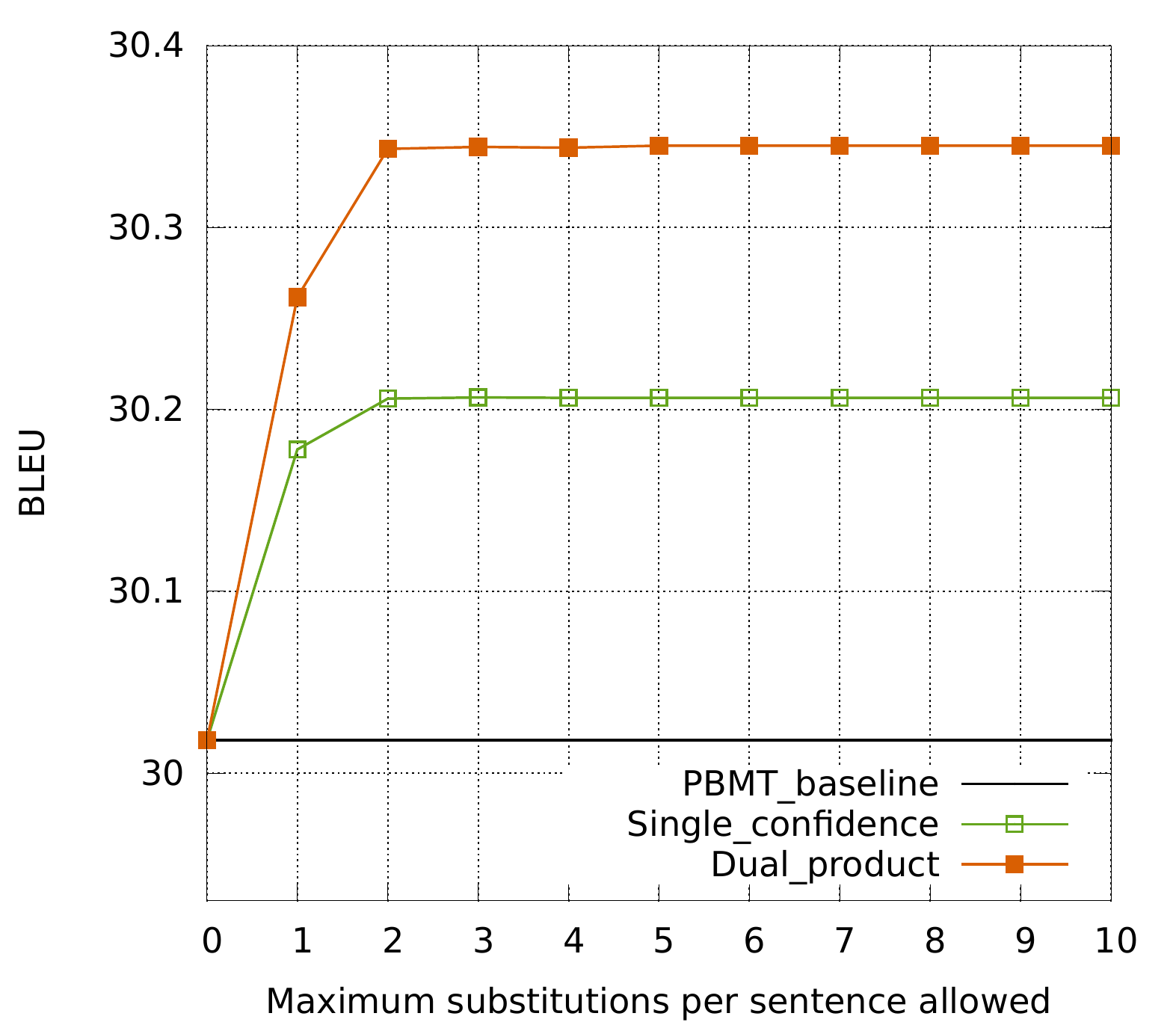}
        \caption{BLEU as a function of the total number of substitutions allowed per sentence. Values are reported on a small $3$K validation set for the single and dual attention models using the best scoring heuristic $\s$ and threshold $t$ (cf. Table~\ref{tab:val_res}).}
        \label{fig:step_to_bleu}
	\end{figure}
\begin{figure}
		\centering
        \includegraphics[width=0.5\textwidth]{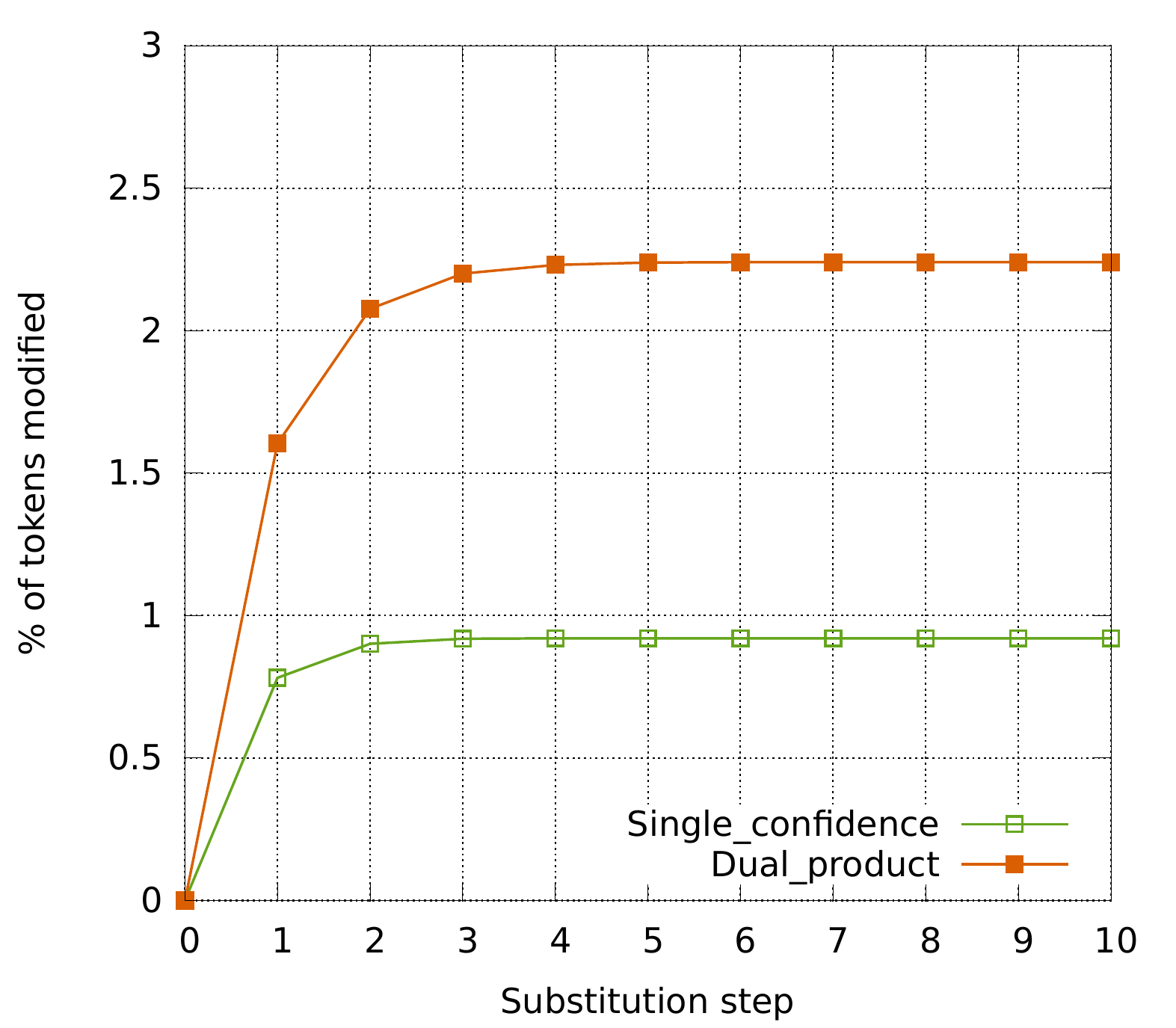}
        \caption{Percentage of modified tokens on the validation set as a function of the total number of substitutions allowed per sentence. All models modify fewer than $2.5\%$ of tokens.}
        \label{fig:step_to_count}
	\end{figure}

Figure \ref{fig:step_to_bleu} plots accuracy versus the number of allowed substitutions and Figure \ref{fig:step_to_count} shows the percentage of actually modified tokens. 
The dual attention model (\textsection\ref{ssec:Dual Attention Model}) outperforms single attention (\textsection\ref{ssec:Attention-based Model}).
Both models achieve most of improvement by making only 1-2 substitutions per sentence. Thereafter only very few substitutions are made with little impact on BLEU.
Figure~\ref{fig:step_to_count} shows that the models saturate quickly, indicating convergence of the refinement output to a state where the models have no more suggestions.
    
To isolate the model contribution from the scoring heuristic, we replace the scoring heuristic with an oracle while keeping the rest of the refinement strategy the same. 
We consider two types of oracle: 
The \emph{full oracle} takes the suggested substitution for each position and then selects which single position should  be edited or whether to stop editing altogether. 
This oracle has the potential to find the largest BLEU improvement. 
The \emph{partial oracle} does not select the position, it just takes the heuristic suggestion for the current step and decides whether to edit or stop the process. 
Notice that both oracles have very limited choice, as they are only able to perform substitutions suggested by our model.

\begin{figure*}
		\centering
        \includegraphics[width=0.48\textwidth]{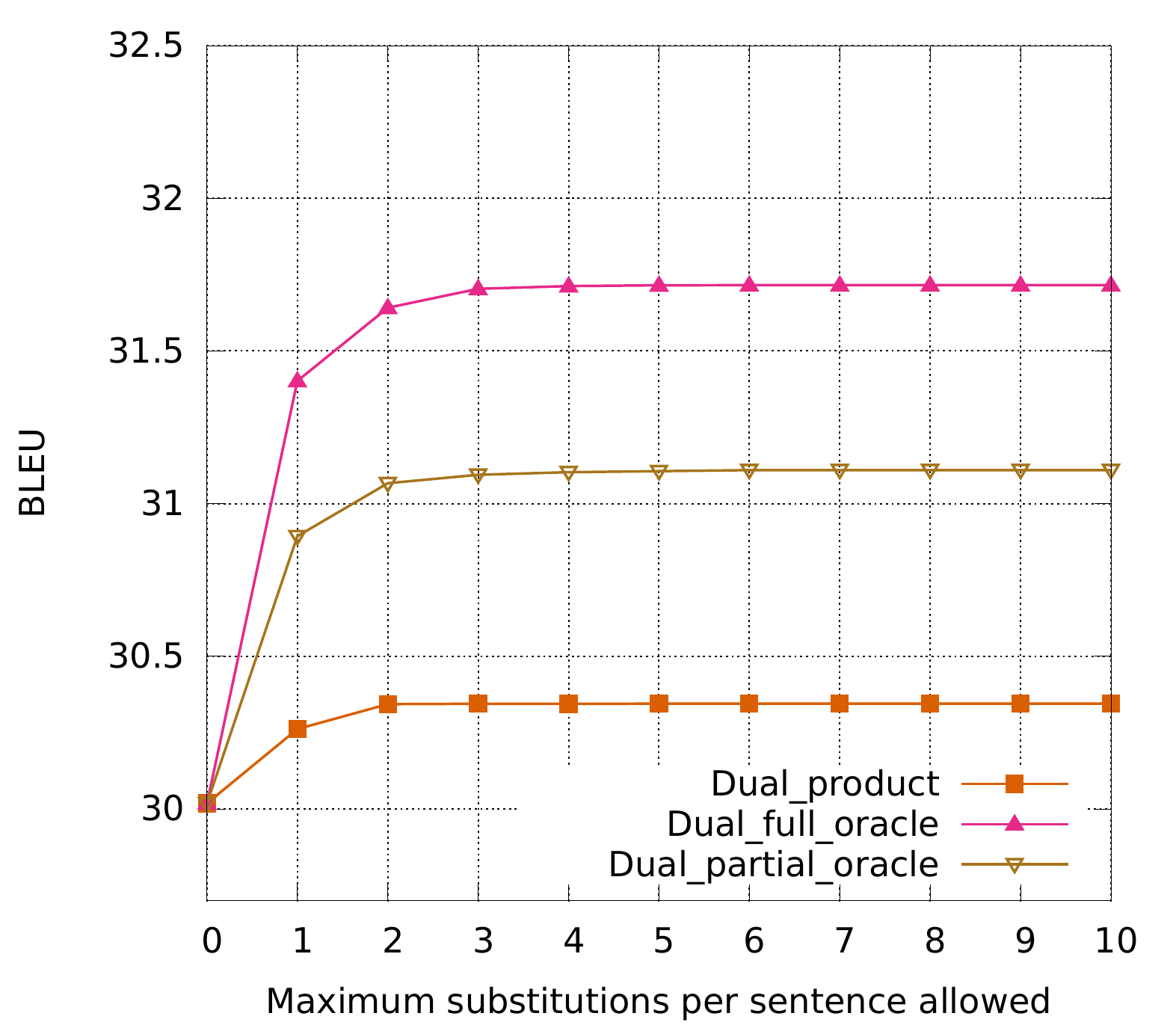}
        \includegraphics[width=0.48\textwidth]{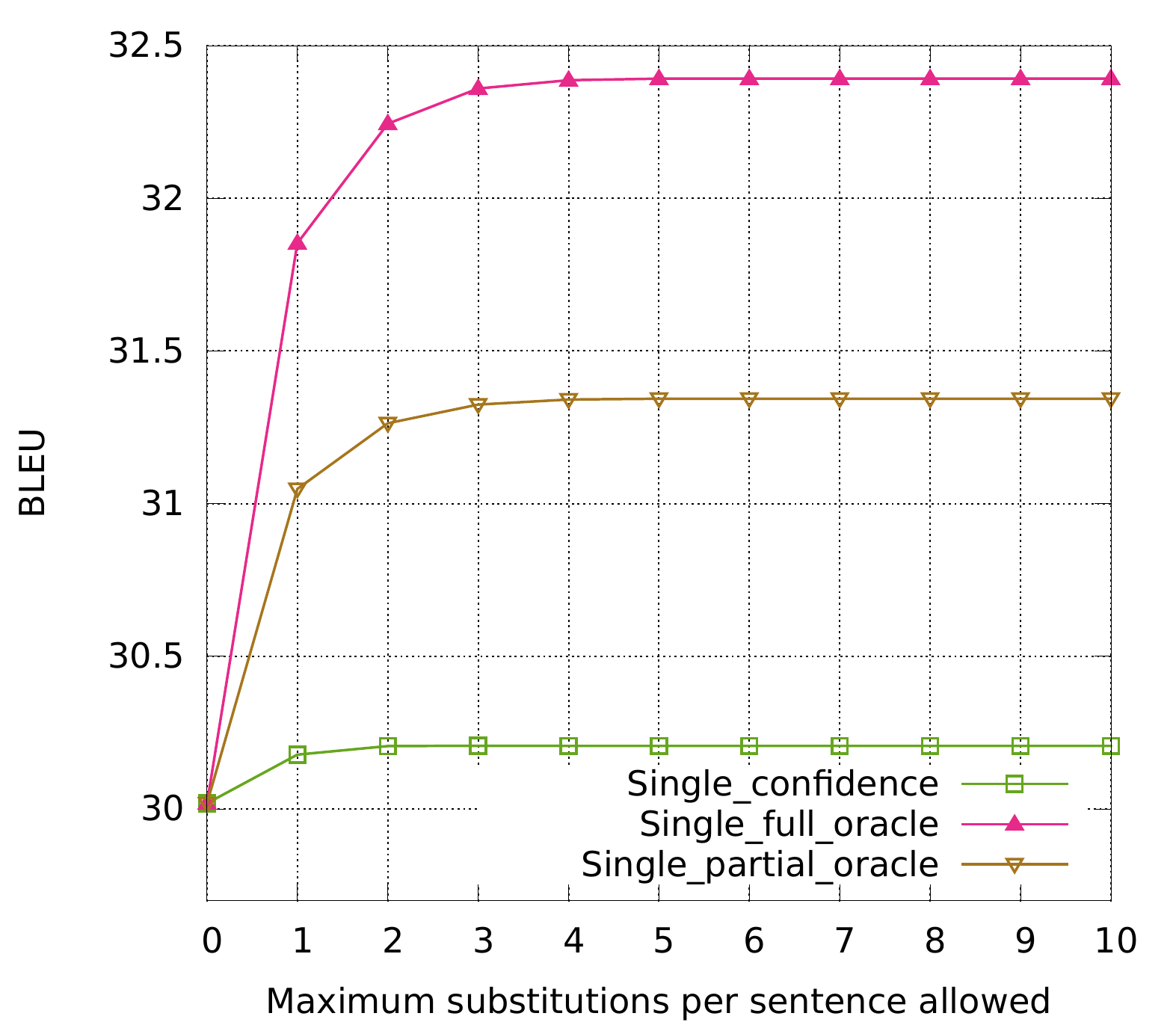}
        \caption{BLEU as a function of the total number of substitutions allowed per sentence. Left: best dual-attention refinement strategy (Dual\_product) versus two oracles. The full oracle (Dual\_full\_oracle) accepts as input $\ypred$ and selects a single $i$ to substitute $\yg^i := \ypred^i$. 
        The partial oracle (Dual\_partial\_oracle) lets the model choose position as well ($i := \arg\max_{1 \leqslant j \leqslant |\yg|}\s(\yg, \ypred)$) but has the ability to prevent substitution $\yg^i := \ypred^i$ if it does not improve BLEU. 
        Right: same for the best single attention setup.}
        \label{fig:step_to_bleu_oracle}
	\end{figure*}

\begin{figure*}
		\centering
        \includegraphics[width=0.48\textwidth]{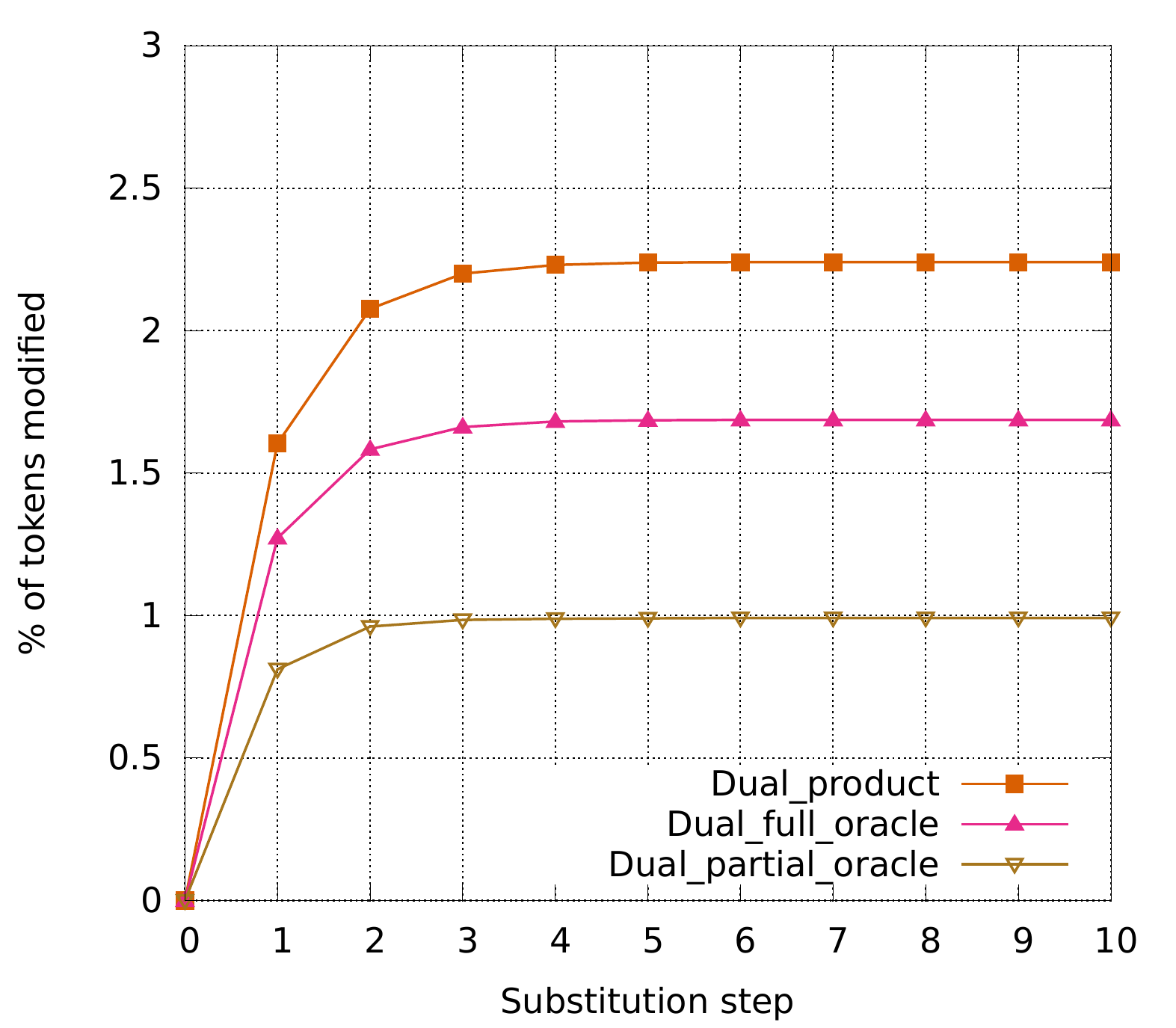}
        \includegraphics[width=0.48\textwidth]{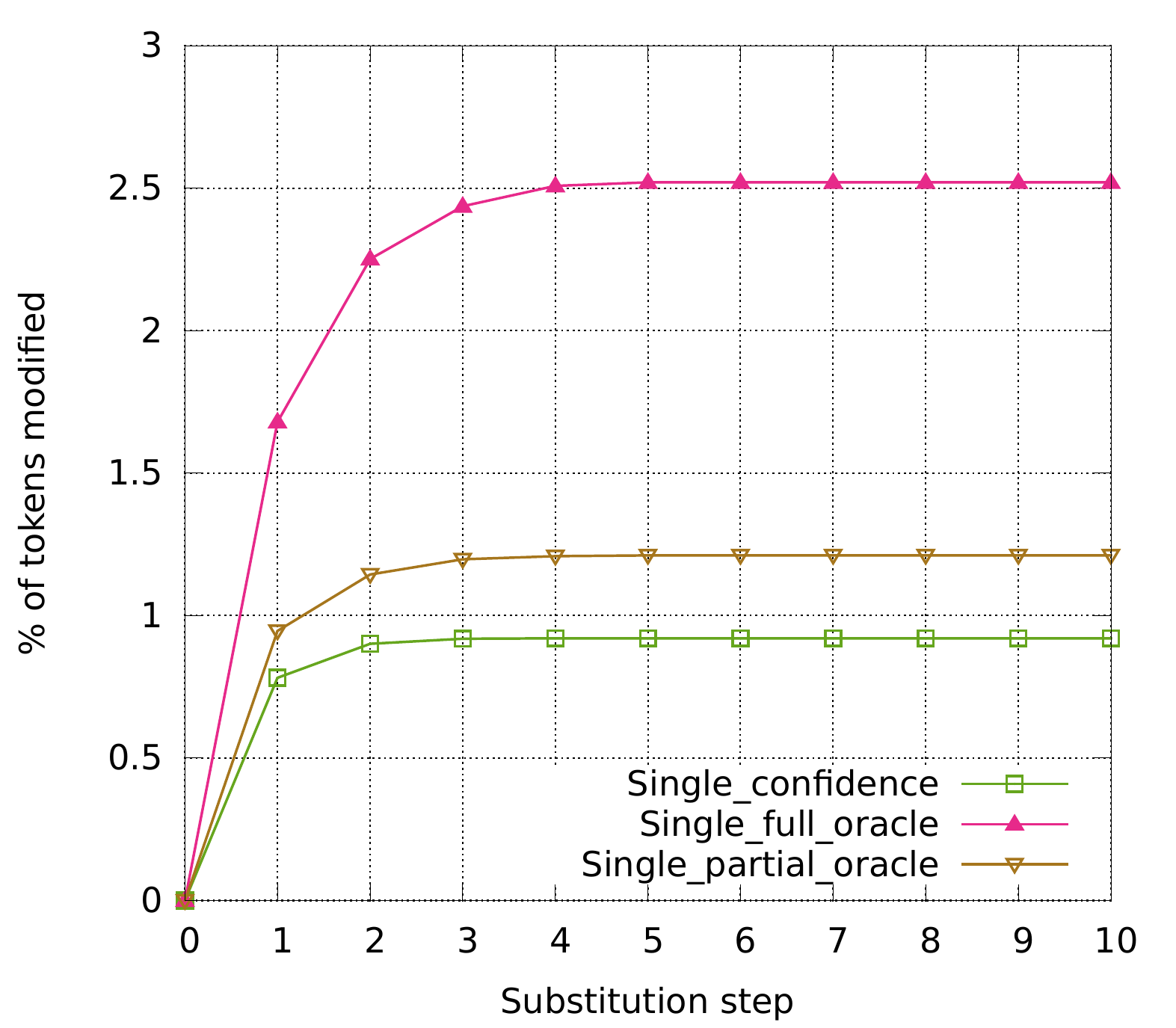}
        \caption{Percentage of modified tokens as a function of total number of substitutions allowed per sentence for the dual attention model (left) and the single attention model (right) compared to the full and partial oracles (cf. Figure~\ref{fig:step_to_bleu_oracle}).}
        \label{fig:step_to_count_oracle}
	\end{figure*}

Figure~\ref{fig:step_to_bleu_oracle} reports the performance of our best single and dual attention models compared to both oracles on the validation set; Figure \ref{fig:step_to_count_oracle} shows the corresponding number of substitutions. The full and partial oracles result in an improvement of $+1.7$ and $+1.09$ BLEU over the baseline in the dual attention setting (compared to $+0.35$ with $\sproduct$).
    
In the single-attention setup the oracles yields a higher improvement ($+2.37$ and $+1.3$) and they also perform more substitutions. This supports our earlier conjecture (\textsection\ref{ssec:Dual Attention Model}) that $\Fdual$ is more conservative and prone to copying words from the guess $\yg$ compared to the single attention model. While helpful in validation, the cautious nature of the dual model restricts the options of the oracle.
    
We make several observations. First, word-prediction models provide high-quality substitutions $\ypred$ that can lead to a significant improvements in BLEU (despite that both oracles are limited in their choice of  $\ypred$). 
This is supported by the simple heuristic $\sconf$ performing very close to more sophisticated strategies (Table~ \ref{tab:val_res}).
    
Second, it is important to have a good confidence estimate on whether a substitution will improve BLEU or not. 
The full oracle, which yields $+1.7$ BLEU, acts as an estimate to having a real-valued confidence measure and replaces the scoring heuristic $\s$. The partial oracle, yielding $+1.09$ BLEU, assesses the benefit of having a binary-valued confidence measure. The latter oracle can only prevent our model from making a BLEU-damaging substitution. However, confidence estimation is a difficult task as we found in \textsection\ref{sec:Error Detection}.
    
Finally, we demonstrate that a significant improvement in BLEU can be achieved through very few substitutions. 
The full and partial oracle modify only 1.69\% and 0.99\% of tokens, or 0.4 and 0.24 modifications per sentence, respectively. 
Of course, oracle substitution assumes access to the reference which is not available at test time. 
At the same time, our oracle is more likely to generate fluent sentences since it only has access to substitutions deemed likely by the model as opposed to an unrestricted oracle that is more likely to suggest improvements leading to unreasonable sentences. 
Note that our oracles only allow substitutions (no deletions or insertions), and only those that raise BLEU in a monotonic fashion, with each single refinement improving the score of the previous translation.

\section{Conclusion and Future Work}\label{sec:Conclusion}

We present a simple iterative decoding scheme for machine translation which is motivated by the inability of existing models to revisit incorrect decoding decisions made in the past. 
Our models improve an initial guess translation via simple word substitutions over several rounds. 
At each round, the model has access to the source as well as the output of the previous round, which is an entire translation of the source. 
This is different to existing decoding algorithms which make predictions based on a limited partial translation and are unable to revisit previous erroneous decoding decisions.
    
Our results increase translation accuracy by up to $0.4$ BLEU on WMT15 German-English translation and modify only $0.6$ words per sentence. 
In our experimental setup we start with the output of a phrase-based translation system but our model is general enough to deal with arbitrary guess translations.
    
We see several future work avenues from here. Experimenting with different initial guess translations such as the output of a neural translation system, or even the result of a simple dictionary-based word-by-word translation scheme. Also one can envision editing a number of guess translations simultaneously by expanding the dual attention mechanism to attend over multiple guesses.

So far we only experimented with word substitution, one may add deletion, insertion or even swaps of single or multi-word units. Finally, the dual-attention model in \textsection\ref{ssec:Dual Attention Model} may present a good starting point for neural multi-source translation \cite{schroeder:2009:eacl}.
    
\section*{Acknowledgments}

We would like to thank Marc'Aurelio Ranzato and Sumit Chopra for helpful discussions related to this work.

\bibliographystyle{eacl2017.bst}
\bibliography{eacl2017.bib}

\appendix

	\begin{table*}[ht]
		\small
		\centering	
        \section{Examples\\[0.2cm]}\label{sec:Examples}
		\begin{tabularx}{0.495\textwidth}{|l|X|}
		\hline
			$\x$    & 	new york city erw\"{a}gt ebenfalls ein solches .\\\hline
			$\yref$ & 	new york city is also considering this .\\\hline
			$\yg$   & 	new york city is also \err{a} \err{such} . \\\hline
			our	    & 	new york city is also \corr{considering} \corr{this} .\\\hline     
		\end{tabularx}
		\begin{tabularx}{0.495\textwidth}{|l|X|}
		\hline
			$\x$    & 	papa , ich bin 22 !\\\hline
			$\yref$ & 	dad , i \&apos;m 22 !\\\hline
			$\yg$   & 	papa , i am 22 \err{.} \\\hline
			our	    & 	papa , i am 22 \corr{!}\\\hline     
		\end{tabularx}
		\\[0.05cm]
		
		\begin{tabularx}{0.495\textwidth}{|l|X|}
		\hline
			$\x$    & 	esme nussbaum senkte ihren kopf .\\\hline
			$\yref$ & 	esme nussbaum lowered her head .\\\hline
			$\yg$   & 	esme nussbaum \err{slashed} \err{its} head .\\\hline
			our 	& 	esme nussbaum \corr{lowered} \corr{her} head .\\\hline     
		\end{tabularx}	
		\begin{tabularx}{0.495\textwidth}{|l|X|}
		\hline
			$\x$    & 	großbritannien importiert 139.000 tonnen .\\\hline
			$\yref$ & 	uk imports 139,000 tons .\\\hline
			$\yg$   & 	britain \err{imported} 139,000 tonnes .\\\hline
			our 	& 	britain \corr{imports} 139,000 tonnes .\\\hline     
		\end{tabularx}
		\\[0.05cm]
				
		\begin{tabularx}{0.995\textwidth}{|l|X|}
		\hline
			$\x$       & alles in deutschland wird subventioniert , von der kohle \"{u}ber autos bis zur landwirtschaft .\\\hline
			$\yref$ & everything is subsidised in germany , from coal , to cars and farmers .\\ \hline
			$\yg$   & \leavevmode\mix{all} in germany \err{,} subsidised by the coal on cars to agriculture .\\ \hline
			$\y$ & \leavevmode\corr{everything} in germany \corr{is} subsidised by the coal on cars to agriculture .\\\hline     
		\end{tabularx}
		\\[0.05cm]
		
		\begin{tabularx}{0.995\textwidth}{|l|X|}
		\hline
			$\x$       & drei m\"{a}nner , die laut aussage der beh\"{o}rden als fahrer arbeiteten , wurden wegen des besitzes und des beabsichtigten verkaufs von marihuana und kokain angeklagt .\\\hline
			$\yref$ & three men who authorities say worked as drivers were charged with possession of marijuana and cocaine with intent to distribute .\\\hline
			$\yg$   & three men who , according to the authorities \err{have} \err{been} \err{worked} as a driver , because of the possession and the \mix{planned} sale of marijuana and cocaine . \\\hline
			$\y$ & three men who , according to the authorities \corr{,} \corr{were} \corr{working} as a driver , because of the possession and the \corr{intended} sale of marijuana and cocaine .\\\hline     
		\end{tabularx}
		\caption{Examples of good refinements performed by our system on our test sets. The model clearly improves the quality of the initial guess translations.\\[0.2cm]}
		\label{tab:good_examples}

		\begin{tabularx}{0.495\textwidth}{|l|X|}
		\hline
			$\x$    & 	er war auch kein klempner .\\\hline
			$\yref$ & 	nor was he a pipe lagger .\\\hline
			$\yg$   & 	he was \mix{also} a plumber .\\\hline
			our 	&	he was \corr{not} a plumber .\\\hline     
		\end{tabularx}
		\begin{tabularx}{0.495\textwidth}{|l|X|}
		\hline
			$\x$    & 	mit 38 aber beging er selbstmord . \\\hline
			$\yref$ & 	but at 38 , he committed suicide . \\\hline
			$\yg$   & 	\leavevmode\err{with} 38 \mix{but} he committed suicide . \\\hline
			our    	& 	\leavevmode\mix{in} 38 \mix{,} he committed suicide . \\\hline
		\end{tabularx}
		\\[0.05cm]
		
		\begin{tabularx}{0.995\textwidth}{|l|X|}
		\hline
			$\x$       & ich habe schon 2,5 millionen in die kampagne gesteckt .\\\hline
			$\yref$ & i have already put 2.5 million into the campaign .\\\hline
			$\yg$   & i have \corr{already} 2.5 million \corr{in} the campaign . \\\hline
			our & i have \corr{put} 2.5 million \corr{into} campaign .\\\hline     
		\end{tabularx}
		\\[0.05cm]
		
		\begin{tabularx}{0.995\textwidth}{|l|X|}
		\hline
			$\x$       & dieses jahr werden amerikaner etwa 106 millionen dollar f\"{u}r k\"{u}rbisse ausgeben , so das us census bureau .\\\hline
			$\yref$ & this year , americans will spend around \$ 106 million on pumpkins , according to the u.s. census bureau .\\\hline
			$\yg$   & this year , the americans \err{are} \corr{approximately} 106 million dollars \corr{for} pumpkins , so the us census bureau .\\\hline
			our & this year , the americans \corr{spend} \corr{about} 106 million dollars \err{to} pumpkins , so the us census bureau .\\\hline     
		\end{tabularx}
		\\[0.05cm]
		
		\begin{tabularx}{0.995\textwidth}{|l|X|}
		\hline
			$\x$       & das thema unterliegt bestimmungen , denen zufolge fluggesellschaften die sicherheit jederzeit aufrechterhalten und passagiere die vom kabinenpersonal gegebenen sicherheitsanweisungen befolgen m\"{u}ssen .\\\hline
			$\yref$ & the issue is covered by regulations which require aircraft operators to ensure safety is maintained at all times and passengers to comply with the safety instructions given by crew members .\\\hline
			$\yg$   & the issue is subject to rules , according to which airlines and passengers \err{to} \err{maintain} \corr{the} security at any time by the cabin crew safety instructions given to follow .\\\hline
			our & the issue is subject to rules , according to which airlines and passengers \corr{must} \corr{follow} \mix{their} security at any time by the cabin crew safety instructions given to follow .\\\hline     
		\end{tabularx}
		\caption{Refinements of mixed quality. Our model is not able to insert new words, and so sometimes it replaces a relevant word with another relevant word. In other cases, improvements are insignificant, or good word replacements are mixed with poor ones.\\[0.2cm]}
		\label{tab:mixed_examples}

		\begin{tabularx}{0.48\textwidth}{|l|X|}
		\hline
			$\x$       & ein krieg , der weder verloren noch gewonnen wird\\\hline
			$\yref$ & a war that is neither lost or won\\\hline
			$\yg$   & a war that is \mix{still} to be \mix{gained} or lost \\\hline
			our & a war that is \mix{neither} to be \err{lost} nor lost\\\hline     
		\end{tabularx}
		\begin{tabularx}{0.51\textwidth}{|l|X|}
		\hline
			$\x$       & werden wir jemals erfahren , was ihn verursacht hat ?\\\hline
			$\yref$ & will we ever know what caused it ?\\\hline
			$\yg$   & will we ever \mix{learn} what caused it ?\\\hline
			our & will we ever \err{hear} what caused it ?\\\hline     
		\end{tabularx}
		\\[0.05cm]
		
		\begin{tabularx}{0.995\textwidth}{|l|X|}
		\hline
			$\x$       & in den vereinigten staaten liegt das durchschnittsalter bei 12,5 jahren , etwas weniger als 12,75 im jahr 1970 .\\\hline
			$\yref$ & in the united states , the average age is 12.5 years , down from 12.75 in 1970 .\\\hline
			$\yg$   & in the united states , the average age \err{at} 12.5 years ago \corr{,} a little less than 12.75 in 1970 . \\\hline
			our & in the united states , the average age \mix{of} 12.5 years ago \err{is} a little less than 12.75 in 1970 .\\\hline     
		\end{tabularx}
		\caption{Examples of poor refinements. Our model does not improve the translation or decreases the quality of the translation.}
		\label{tab:bad_examples}
	\end{table*}

\end{document}